# Analysis of Visual Question Answering Algorithms with attention model

Param Ahir[a*], Hiteishi M. Diwanji[a]

[a]L. D. Enginerring College, Ahmedabad, India

,______________________________________________________________________________

**Abstract**

Visual question answering (VQA) usesimage processing algorithms to process the image and natural language processing methods to understand the question and answer it. VQA is helpful to a visually impaired person, can be used for the security surveillance system and online chatbots that learn from the web. It uses NLP methods to learn the semantic of question and to derive the textual features. Computer vision techniques are used for generating image representation in such a way that they can identify the objects about which question is asked. The Attention model tries to mimic the human behavior of giving attention to a different region of an image according to our understanding of its context. This paper critically examines and reviews methods of VQA algorithm such as generation of semantics of text, identification of objects and answer classification techniques that use the co-attention approach.

______________________________________________________________________________

## 1. Introduction

Visual question answering system can help in humanizing human-computer interactions in the artificial intelligence field in such a way that it becomes similar to human conversations. It is a multi-disciplinary research problem and requires concurrent processing of textual features from a question and visual features from the image It uses NLP to understand the input questionand answer it. It is significantly different from the normal NLP problem as it requires analysis and reasoning of text over the content of the image. Object recognition techniques help in identifying the content of the image. To make the process simpler one can derive which areas of an image are important to answer the given question by providing those parts of the question to the image processing module. So that it gives attention to only essential regions of an image and process them only. In VQA system text analysis and image analysis are mutually dependent on each other.As a human, we can easily identify objects, their position and surrounding in an image, understand the question and its relation to image and can use the knowledge and common sense to answer it. When we want a computer system to perform the same tasks systematic approach and algorithms are required. The process of the VQA system contains three modules, (i) Question features extraction (ii) Image feature extraction (iii) Answering Module. Various deep learning techniques are used to implement these modules. Forprocessing and extraction of text features recurrent neural network (RNN) is used. For processing and extraction of image features convolution neural network (CNN) is used. To predict the correct answer various classification methods are used.

## 2. Basic Concept

Earlier basic baseline models were used to answer the question about the image. Those models answer the question by giving the most frequent answers. Some models even answer the question by randomly picking the answer and then checking its accuracy with various loss functions. Later on, some sophisticated models with a linear classifier or multilayer perceptron were used. Vector representation of the combination of textual and image features are given as input to the multilayer perceptron. Various methods were used to combine these features like simple concatenation, sum pooling, average pooling or product of features, etc. Most of the previous works deal with two models. First model Simple multilayer perceptron (MLP) [1] used a neural network classifier with two hidden layers. Image features combined with textual features were given as input. To derive the output tanh activation function is used. For textual features representation, a bag-of-words method was used. For image features, the output of the last layer of ResNet (visual geometry group) was used. Second model long short-term memory (LSTM) [2] used one-hot encoding for question features and for, image features are derived just like MLP but features are transformed into a linear vector of 1024 dimension to match it with the question feature vector. The basic problem with using global features is that it generates obscure input space for the model. It is important to attend the most relevant region of the input space to get clarity about its target area that should be looked upon to generate the answer. An issue with these models is that they include global image features in processing and generation of the answer. Contrary to that attention model only focuses on local features of the image which are derived using the textual attention model.

______________________________

*Param Ahir
*E-mail address:* ahirparam@gmail.com

## 3. Related Work

In VQA other than the baseline model, there are various algorithms available for different models. There are various methods to retrieve features of image and text, to represent those features and fuse them. In the Bayesian model, one method uses semantic segmentation to identify objects and their position. Another algorithm use answer type with question feature to predict the probability of image features. In Multimodal compact bilinear (MCB) pooling, the fusion matrix is created using the outer-product of low dimension feature matrix. There is one more technique that uses Hadamard product and linear mapping to fuse the features called multi-modal low-rank bilinear pooling (MLB). In VQA all questions are not unassuming. Certain questions required more than one step to understand and answer. These sub-tasks can be processed separately. The Neural Module Network (NMN) uses special question parser to find out such sub-tasks and process them individually and combine later. In VQA some questions can't be answered just by deriving image features. First, comprehend the question in depth, which required some external knowledge about the context of the question. There are models available that prove such an external knowledge base in the system. Dynamic Parameter Prediction Model makes sure that image features are completely related to the question dynamically-generated question parameters are added into the CNN which are generated using a recurrent neural network. This mechanism provides implicit attention to image features so that input image features are related to question only. Another method Multimodal residual network derives visual and textual features using residual mapping. Embedding of question and visual features are fused using different techniques. The Attention model contains vectors that assign weights to the regions of image and words of a sentence. A similar approach is used for the sentence where a specific word can increase the probability of occurrence of another word. There are many mechanisms available to implement attention like, dot-product, scale dot-product, content-base and location base. Broadly attention mechanism can be divided into three categories, Self-attention – Try to find a relation between input sequences by relating words at a different position in input sequence in each trial. Global-attention – Attend entire input sequence .Local Attention – Attend parts of the input sequence. Such several VQA models are described in the following section.

## 4. Analysis of Various VQA Algorithms that use Attention Model

*4.1 Beyond Bilinear: Generalized Multimodal Factorized High-Order Pooling for Visual Question Answering [3]*

In this paper, two models are given, first Baseline Model extracts the features of the image using a residual neural network with 152 layers. Image is represented using the 2048 dimension features generated by the output of pool5. Questions are first to tokenize into words and then one-hot encoding is performed and one hot vector containing features is derived. These vectors are given as input in long short term memory (LSTM) recurrent neural network with 1024 hidden layers. The Output feature of the last word from the LSTM network is used to represent the question. Multimodal features that are extracted from image and text are fused using MFB or MFH and output feature z is generated. For answer prediction they take N most frequent answers as N classes and fed z into N-way classifier. Second Coattention Model uses similar approach to derive image and text features and then output features of questions are given into question attention module which generates attentive question representation. This output is given to image attention module and using MFH, fusion of image features and text features are generated. In both image and question attention module convolutional layers of $1 \times 1$ and ReLU layers are given. Its output is given to softmax normalization layers to predict the attention weight for each input feature. The attentive feature is obtained by the weighted sum of the input features. Question attention module is learning in a self-attentive manner using the question feature to make question explicable without image.

It uses the MFB model to provide a fusion of textual and image features. It uses a generalized MFH approach to cascade multiple MFB to find complex correlations to represent an accurate distinction between different question-image pairs. In Multimodal Factorized Bilinear Pooling (MFB) if we consider visual feature $x \, \varepsilon \, R^m$ for image and textual feature $y \, \varepsilon \, R^n$ for question then simplest multimodal bilinear model is $Z_i = x^T W_i y$ where $W_i \, \varepsilon \, R^{m \times n}$ is projection matrix and $Z_i$ is the output of the bilinear model. Two lower-level projection matrices are created using matrix factorization. To reduce dimensionality and generate output feature sum pooling is applied on U and V vector. Multimodal factorized high-order pooling (MFH) model derives its output feature Z by concatenating output features of multiple MFB blocks which are derived by performing expansion and squeezing as shown above. $Z= MFH_p = [z_1, z_2,. . . z_p]$ P<4 (P is the number of MFB blocks).In model image and question is given as input and feature extraction on both is performed and then features of both are integrated using MFB and MFH. Each derived answer is considered as one class of answer and multiclass classification is performed to predict the correct answer. In ANSWER CORRELATION MODELING each answer is assigned a weight and presented with their probability distribution. KLD (Kullback-Leibler divergence) loss function is applied to it, to measure the dissimilarity among two probability distributions.

*4.2 Bottom-Up and Top-Down Attention for Image Captioning and Visual Question Answering [4]*

This model determines the image regions and generates feature vectors using bottom-up approach and for weighting each feature and image captioning top-down approach is used. Bottom-up attention model uses faster R-CNN algorithm that detects objects in two stages. The first stage is the Region Proposal Network (RPN) finds out the regions that possibly contains the object. To find out the non-object based segmentation it considers image features such as color similarity, textual similarity, region size, and region filling, then it merges together such smaller regions to generate larger areas and finds out which of these anchors are foreground and background and using training dataset label objects. There are the different sizes of regions available that are reduced to the same size using ROI Pooling. To train model initiated faster R-CNN with ResNet and embed ground truth objects with mean pool convolution features is used. Captioning Model uses top-down attention mechanism where LSTM levels are defined; the first layer is top-down visual attention layer and second is language model. The output of bottom-up attention model is taken as features V and input vector for each attention LSTM step is the previous output of language LSTM step and mean pool features vector. In Language LSTM just like attention input is the previous output of attention LSTM and image feature. Its VQA Model encodes each question as to the hidden state q of a gated recurrent unit and represents it using word embedding. The joint representation of image and question is provided. Learning is done using Self-critical sequence training (SCST) algorithm.

*4.3 Co-Attention Network with Question Type for Visual Question Answering [5]*

The objective of model is to add question type into the co-attention mechanism to decrease the number of possible answer and make process of selecting answer relatively fast. The main three modules of model are input representation module, co-attention module and question type module. In Input Representation image embedding and question embedding is implemented. In Image Embedding objects in image are derived using a bottom-up approach and faster R-CNN algorithm. In Question Embedding question is tokenized into words and one hot features vector is generated. It is given as input in Bi-LSTM, iteration expresses question at each hidden layer. In Co-attention module there are two sub-modules. In Textual attention based on self-attention, it generates 2-D matrix that represents sentence embedding of a question and contains all or most of the part of the question. Each input vector in self-attention mechanism with weights for all the words. Output of this module is sentence with highlighted. In Question guided visual attention question representation M is converted into m' by applying mean function. Question representation m' and image vector V is projected in the same dimension and multiplication wise fusion is applied. Using linear layer and softmax normalized attention to the weight of each image object is generated, which is given in a single vector. In Question Type module many question types like color, time, counting, location, reason, sport, judgment etc. are used. One hot encoding of these question type is embedded with multimodal feature fusion and given to classifier for answer generation. In Prediction and Learning phase set of candidate answers are given into multi-layer perceptron to predict the correct answer, the sigmoid activation function is used to predict the score of each answer.

*4.4 Cross-Modal Multistep Fusion Network with Co-Attention for Visual Question Answering [6]*

AIM is to use cross modal multistep fusion in place of simple multi modal fusion. In Co-attention module sentence-guided word attention and question-guided visual attention is used. Answer prediction uses KLD loss function (Kullback-Leibler Divergence).For Sentence-guided word attention context of the question is taken into consideration for deciding importance of individual words. Attention score of each word is transformed into expectation of word. Expectation $P_i$ = softmax(conv(mlb($w_i$,s))). $W_i$ is sentence feature derived from last hidden layer of LSTM and word features derived in each time state of LSTM. mlb() is multimodal low rank bilinear pooling and conv() is convolution learning operation that consist of 1X1 of convolution layers and ReLu layers. Output here is $\bar{W}$ word attention. In Question-guided visual attention

$$\bar{X} = \sum_{i=1}^{M} Softmax(conv(mlb(x_i, \bar{W}))) \bullet x_i$$

Where X is the image feature extracted by deep residual network. In CROSS-MODAL MULTISTEP FUSION NETWORK output from image and word attention is fed as input into CMF network then at each layer three outputs are generated. Out of them two attention features are given for next CMF unit and fusion feature provides multistep fusion using sum pooling to get the final feature for answer prediction. This fusion feature is generated by hadamard product between two models and then output is standardized by power normalization and L2 normalization. For Answer Prediction Kulluback-Leibler Divergence (KLD) loss using label distribution learning with a fixed answer distribution is used.

*4.5 Object-Difference Attention A Simple Relational Attention for Visual Question Answering [7]*

Aim is to develop a VQA system where attention mechanism gives different attention to different objects of an image. For Data Embedding faster R-CNN is used to encode image features using bottom-up attention and GRU is used for encoding text features. In Object Difference Attention Comparison between two objects in image is guided by the question using softmax activation function. In Decision Making ODA model is called multiple time with the learning parameter to obtain more information about the attention region. Image and question features are fused to get candidate answers and sigmoid activation function is applied to get the predicted score of each candidate answer. The model is trained using KLD loss function.

*4.6 Visual Question Answering using Explicit Visual Attention [8]*

The aim is to use pictorial superiority theory in image attention model of the VQA system.Previous triplet based classification model has implicit attention mechanism for image attention which sometimes could not find important object regions effectively. Explicit Attention Model bridges the semantic gap between the textual and image representation. Feature map from the question representation is generated by applying the convolution neural network. Feature map with the word embedding of the question is given into multilayer perceptron with hidden layers to generate image representation that has image attention regions. To train this attention model ground truth bounding boxes assigned to feature map using nearest neighbour interpolation are used. There are two modules in VQA model, feature embedding layer and multiple choice layers. In Feature Embedding Layer representations from question and image is extracted. Attention vector is derived using explicit attention model. And embedding vectors are generated to encode the question and answer, but unlike previous models here for textual visual tasks separate model is used. For question features $Q_f$ and for answer features $A_f$ notations are used. For Multiple Choice Layer triplet of question-image-answer is given as input in MLP layer to predict whether the triplet is correct or not and it will give scalar values as an output that indicates the correctness of given triplet. Extracted feature vectors are not injected directly instead of similarity and distance between representations of image, question and answer is calculated and that is fed into the classifier. So vector that fed into classifier is [ $Q_f$ ;$A_f$; $Q_f \odot A_f$ ; $\| Q_f - A_f \|$ ; $I_{m'}$ ; $I_{m'} \odot Z$ ] $\odot$ is Hadamard product ,$I_{m'}$ is attention representation vector derived from attention model. Z is the output of the transformation layer that generates common representation for the concatenated vector of question and answer. This is computed as, $T_{qa} = [ Q_f, A_f ] \in R^{2D_w}$, $Z^{(n)} = \sigma ( t_{qa} W_{qa}^{(n)} + b_{qa}^{(n)} ) \in R^{D_d}$ , $W_{qa}^{(n)}$ and $b_{qa}^{(n)}$ are parameters of transformation layer. $\sigma( )$ is sigmoid activation function. To computing this vector MLP with 8096 hidden units with rectifier activation function is used and for the final layer sigmoid activation function is used. Binary logistic loss function is used for optimization.

## 5. Conclusion

This paper presents an analytical assessment of feature representation and fusion practices used in Visual question answering. Examination of mutual methodologies for merging convolutional and recurrent neural networks to fuse images and questions to a conjoint feature space is performed. Various popular attention models are analyzed against the requirements of VQA. There are some limitations of these algorithms which points to future directions in this field. Like in [3] if the numbers of MFB blocks aremore than 4 to 5 then what will happen. In [4] if a select small number of objects from the given feature map in the last convolution layer and what if sum pooling or max pooling is used instead of mean pooling. In [5] one more question type named 'absurd' can be added which tells whether the question is answerable or not also add the knowledge of question type at the time when applying attention algorithm to an image. In [6] what if complex fusion method is combined with multiple fusion networks. In [7] can use linguistic reasoning for ODA model. In [8] what to do with a question that does not contains ground truth boxes. Use the hybrid implicit – explicit model and techniques like bag-of-features can be used for textual representation. Knowledge base can be added to this model in such a way that it does not over train the model and does not answer questions about objects not present in image. These future works can help improve the VQA system and benefits the artificial intelligence field in general.